%% file: main.tex
\documentclass[10pt,twocolumn,letterpaper]{article}

\usepackage{cvpr}              

\makeatletter
\@namedef{ver@everyshi.sty}{}
\makeatother

\usepackage{graphicx}
\usepackage{amsmath}
\usepackage{amssymb}
\usepackage{booktabs}
\usepackage{import}
\usepackage{svg}

\usepackage[pagebackref,breaklinks,colorlinks]{hyperref}

\usepackage[capitalize]{cleveref}
\crefname{section}{Sec.}{Secs.}
\Crefname{section}{Section}{Sections}
\Crefname{table}{Table}{Tables}
\crefname{table}{Tab.}{Tabs.}

\def\cvprPaperID{9757} 
\def\confName{CVPR}
\def\confYear{2023}

\def\conv{\ast}

\begin{document}

\definecolor{MyDarkBlue}{rgb}{0,0.08,1}
\definecolor{MyCyan}{rgb}{0,0.6,0.6}
\definecolor{MyDarkGreen}{rgb}{0.02,0.6,0.02}
\definecolor{MyDarkRed}{rgb}{0.8,0.02,0.02}
\definecolor{MyDarkOrange}{rgb}{0.40,0.2,0.02}
\definecolor{MyPurple}{RGB}{111,0,255}
\definecolor{MyRed}{rgb}{1.0,0.0,0.0}
\definecolor{MyGold}{rgb}{0.75,0.6,0.12}
\definecolor{MyDarkgray}{rgb}{0.66, 0.66, 0.66}
\definecolor{MyOrange}{rgb}{1.0, 0.6, 0.0}
\definecolor{MyGreen}{rgb}{0.02,0.5,0.02}

\newcommand{\sean}[1]{\textcolor{MyPurple}{[Sean: #1]}}
\newcommand{\david}[1]{\textcolor{MyGreen}{[David: #1]}}
\newcommand{\todo}[1]{\textcolor{MyRed}{[TODO: #1]}}
\newcommand{\rohit}[1]{\textcolor{MyOrange}{[Rohit: #1]}}
\newcommand{\vecore}[1]{\textcolor{MyCyan}{[Vecore: #1]}}
\newcommand{\brian}[1]{\textcolor{MyDarkRed}{[Brian: #1]}}

\title{Controllable Light Diffusion for Portraits\\}

\author{
David Futschik$^{1,2}$
\quad
Kelvin Ritland$^{1}$
\quad
James Vecore$^{1}$
\quad
Sean Fanello$^{1}$ \\
\quad
Sergio Orts-Escolano$^{1}$
\quad
Brian Curless$^{1,3}$
\quad
Daniel S\'{y}kora$^{1,2}$
\quad
Rohit Pandey$^{1}$\\ \\
\quad $^{1}$Google Research
\quad $^{2}$CTU in Prague, FEE
\quad $^{3}$University of Washington
}

\twocolumn[{%
\renewcommand\twocolumn[1][]{#1}%
\vspace{-1.6em}
\maketitle
\vspace{-1.5em}

\begin{center}
    \centering
    \includegraphics[width=0.98\linewidth, trim={0 0 0 0}, clip]{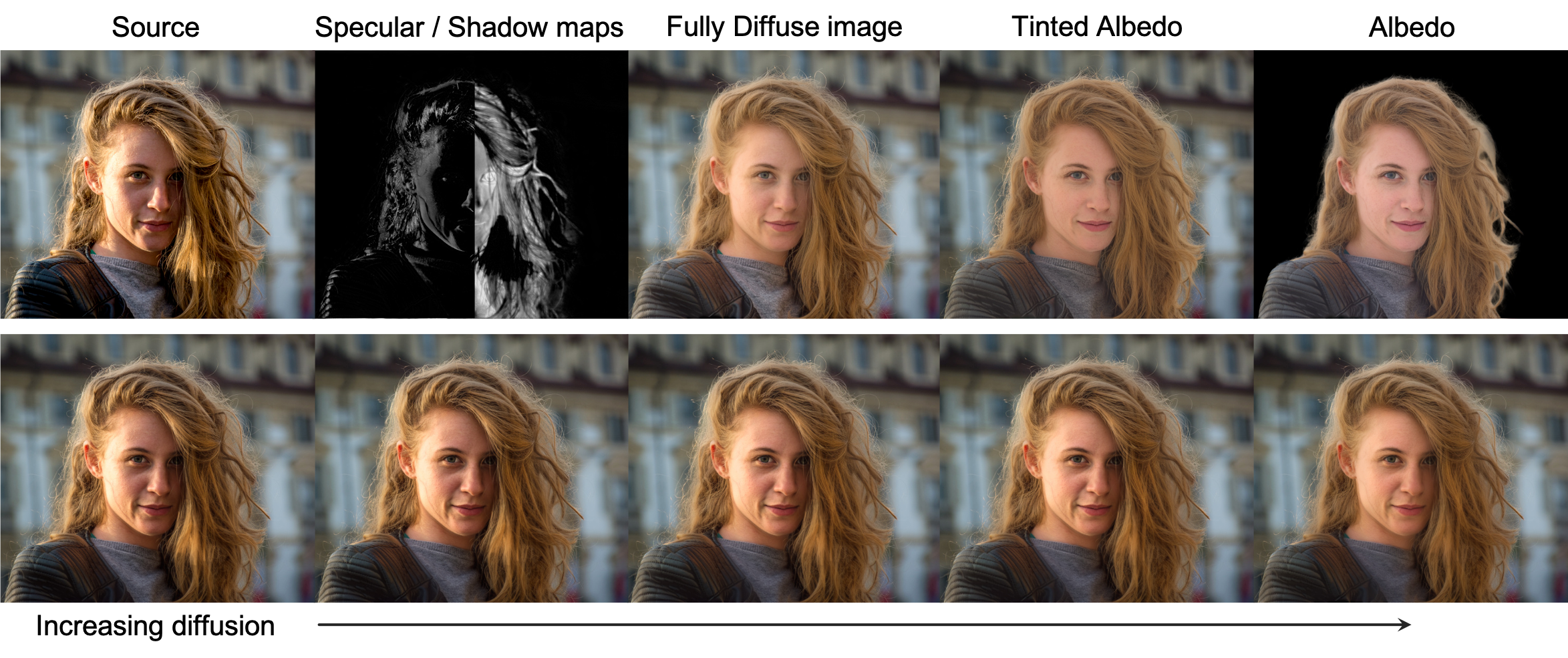}
    \captionof{figure}{Our method enables control over the diffuseness of light in arbitrary portrait images. We first extract specular/shadow maps from the input image and then produce a fully diffuse image. Additionally, we show how to recover a uniformly lit image, \ie tinted by the average light color, from which we can estimate the untinted albedo.  The bottom row illustrates the application of editing the input photo by gradually increasing the amount of light diffusion.}

    \vspace{-0.0em}
    \label{fig:teaser}
\end{center}%
}]



\begin{abstract}
\vspace{-4.2mm}
We introduce {\rm light diffusion}, a novel method to improve lighting in portraits, softening harsh shadows and specular highlights while preserving overall scene illumination. Inspired by professional photographers' diffusers and scrims, our method softens lighting given only a single portrait photo. Previous portrait relighting approaches focus on changing the entire lighting environment, removing shadows (ignoring strong specular highlights), or removing shading entirely. In contrast, we propose a learning based method that allows us to control the amount of light diffusion and apply it on in-the-wild portraits. Additionally, we design a method to synthetically generate plausible external shadows with sub-surface scattering effects while conforming to the shape of the subject's face. Finally, we show how our approach can increase the robustness of higher level vision applications, such as albedo estimation, geometry estimation and semantic segmentation.
\end{abstract}

\input{introduction.tex}
\input{related_work.tex}
\input{method.tex}
\input{experiments.tex}
\input{conclusion.tex}

\section*{Acknowledgements}

David Futschik and Daniel S\'ykora were partly supported by the Research Center for Informatics, grant No.~CZ.02.1.01/0.0/0.0/16\_019/0000765 and by the Grant Agency of the Czech Technical University in Prague, grant No.~SGS22/173/OHK3/3T/13.

{\small
\bibliographystyle{ieee_fullname}
\bibliography{egbib}
}

\end{document}

%% file: introduction.tex
\section{Introduction}
\label{sec:intro}

\begin{figure}
    \centering
    \includegraphics[width=\linewidth]{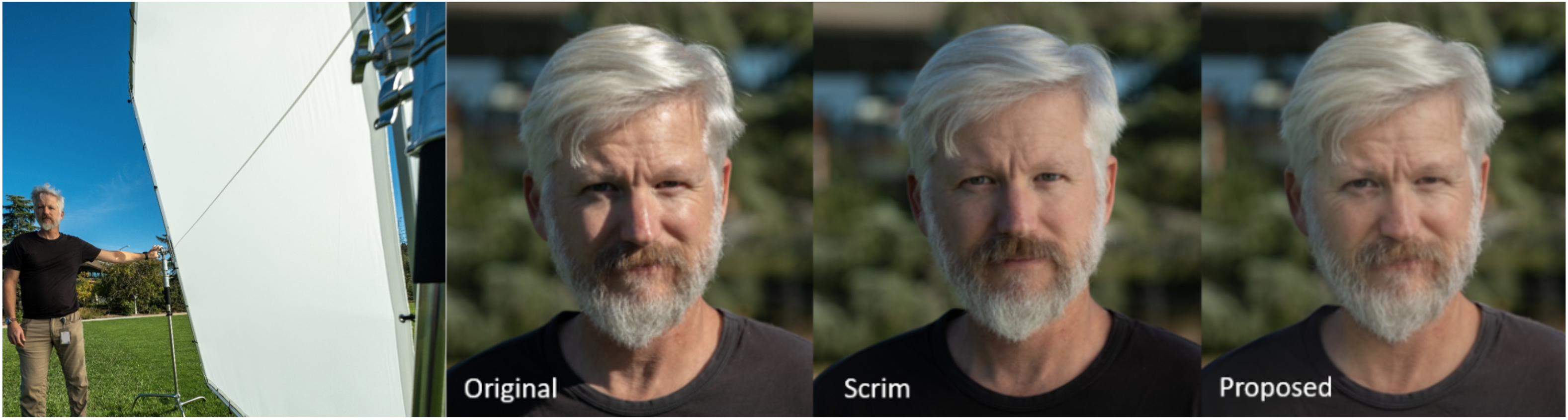}
    \caption{Using a bulky scrim (left), a photographer can reduce strong shadows and specularities.  Our proposed approach operates directly on the original image to produce a similar softening effect.}
    \label{fig:scrim_comparison}
\end{figure}

High quality lighting of a subject is essential for capturing beautiful portraits. Professional photographers go to great lengths and cost to control lighting. Outside the studio, natural lighting can be particularly harsh due to direct sunlight, resulting in strong shadows and pronounced specular effects across a subject’s face.  While the effect can be dramatic, it is usually not the desired look.  Professional photographers often address this problem with a scrim or diffuser (Figure \ref{fig:scrim_comparison}), mounted on a rig along the line of sight from the sun to soften the shadows and specular highlights, leading to much more pleasing portraits~\cite{grey04}.  Casual photographers, however, generally lack the equipment, expertise, or even the desire to spend time in the moment to perfect the lighting in this way.  We take inspiration from professional photography and propose to diffuse the lighting on a subject in an image, i.e., directly estimating the appearance of the person as though the lighting had been softer, enabling anyone to improve the lighting in their photos after the shot is taken.  

Deep learning approaches have led to great advances in relighting portraits~\cite{sun2019single, zhou2019portraitrelighting, nestmeyer2020learning, shu2017portrait, Wang20, zhang2020portrait,pandey2021total,zhang21,ji2022relight,yeh2022learning}.  Our goal is different: we want to improve the existing lighting rather than replace it entirely.  This goal has two advantages: the resulting portrait has improved lighting that is visually consistent with the existing background, and the task is ultimately simpler, leading to a more robust solution than completely relighting the subject under arbitrary illumination.  Indeed, one could estimate the lighting \cite{LeGendre20,LeGendre19}, diffuse (blur) it, and then relight the subject \cite{pandey2021total,ji2022relight,yeh2022learning}, but lighting estimation and the full relighting task themselves are open research questions.  We instead go directly from input image to diffused-lighting image without any illumination estimation.

Past works~\cite{zhang2020portrait,inoue21} specifically focused on removing shadows from a subject via CNNs.  However, these methods do not address the unflattering specularities that remain which our work tackles.

In the extreme, lighting can be diffused until it is completely uniform.  The problem of “delighting,” recovering the underlying texture (albedo) as though a subject has been uniformly lit by white light\footnote{Technically, uniform lighting will leave ambient occlusion in the recovered albedo, often desirable for downstream rendering tasks.}, has also been studied (most recently in~\cite{Weir2022}). The resulting portrait is not suitable as an end result – too flat, not visually consistent with the background – but the albedo map can be used as a step in portrait relighting systems~\cite{pandey2021total}.  Delighting, however, has proved to be a challenging task to do well, as the space of materials, particularly clothing, can be too large to handle effectively.

In this paper, we propose {\em light diffusion}, a learning-based approach to controllably adjust the levels of diffuse lighting on a portrait subject.  The proposed method is able to soften specularities, self shadows, and external shadows while  maintaining  the  color  tones  of  the  subject, leading to a result that naturally blends into the original scene (see Fig. \ref{fig:teaser}). Our variable diffusion formulation allows us to go from subtle shading adjustment all the way to removing the shading on the subject entirely to obtain an albedo robust to shadows and clothing variation.


Our overall contributions are the following:
\begin{itemize}
    \item A novel, learning-based formulation for the light diffusion problem, which enables controlling the strength of shadows and specular highlights in portraits.
    \item A synthetic external shadow generation approach that conforms to the shape of the subject and matches the diffuseness of the illumination.
    \item A robust albedo predictor, able to deal with color ambiguities in clothing with widely varying materials and colors. 
    \item Extensive experiments and comparisons with state-of-art approaches, as well as results on downstream applications showing how light diffusion can improve the performance of a variety of computer visions tasks.
\end{itemize}



%% file: related_work.tex
\section{Related Work}
\label{sec:related_work}
Controlling the illumination in captured photos has been exhaustively studied in the context of portrait relighting \cite{sun2019single, zhou2019portraitrelighting, nestmeyer2020learning, shu2017portrait, Wang20, zhang2020portrait,pandey2021total,zhang21,ji2022relight,yeh2022learning}, which tries to address this problem for consumer photography using deep learning. Generative models and inverse rendering \cite{tan2022volux,pan2021shadegan,Abdal21,deng2020disentangled,mallikarjun2021} have also been proposed to enable face editing and synthesis of portraits under any desired illumination.

The method of Sun et al. \cite{sun2019single}, was the first to propose a self-supervised architecture to infer the current lighting condition and replace it with any desired illumination to obtain newly relit images. This was the first deep learning method applied to this specific topic, overcoming issues of previous approaches such as \cite{shu2017portrait}. 

However, this approach does not explicitly perform any image decomposition, relying on a full end-to-end method, which makes its explainability harder. More recent methods \cite{Wang20,nestmeyer2020learning,pandey2021total,ji2022relight} decompose the relighting problem into multiple stages. These approaches usually rely on a geometry network to predict surface normals of the subject, and an albedo predictor generates a \textit{de-lit} image of the portrait, that is close to the actual albedo (i.e. if the person was illuminated by a white diffuse light from any direction). A final neural renderer module combines geometry, albedo and target illumination to generate the relit image. Differently from previous work, Pandey et al. \cite{pandey2021total} showed the importance of a per-pixel aligned lighting representation to better exploit U-Net architectures \cite{unet}, showing state-of-art results for relighting and compositing.

Other methods specifically focus on the problem of image decomposition \cite{Abdelrehim07,BarronM15,meka:2017,Ren:2015,Meka:2018,xu2018deep,Kanamori:2018,Weir2022,yang22}, attempting to decompose the image into albedo, geometry and reflectance components. Early methods rely on model fitting and parametric techniques \cite{blanz99,Abdelrehim07,BarronM15,meka:2017}, which are limited in capturing high frequency details not captured by these models, whereas more recent approaches employ learned based strategies \cite{Ren:2015,Meka:2018,xu2018deep,Kanamori:2018}. 

In particular, the method of Weir et al. \cite{Weir2022} explicitly tackles the problem of \textit{portrait de-lighting}. This approach relies on novel loss functions specifically targeting shadows and specular highlights. The final inferred result is an albedo image, which resembles the portrait as if it was lit from diffuse uniform illumination. Similarly, Yang et al. \cite{yang22} propose an architecture to remove facial make-up, generating albedo images.

These methods, however completely remove the lighting from the current scene, whereas in photography applications one may simply want to control the effect of the current illumination, perhaps softening shadows and specular highlights. Along these lines the methods of Zhang et al. \cite{zhang2020portrait} and Inouei and Yamasaki \cite{inoue21} propose novel approaches to generate synthetic shadows that can be applied to in-the-wild images. Given these synthetically generated datasets, they propose a CNN based architecture to learn to remove shadows. The final systems are capable of removing harsh shadows while softening the overall look.  Despite their impressive results, these approaches are designed to deal with shadow removal, and, although some softening effect can be obtained as byproduct of the method, their formulations ignore high order light transport effects such as specular highlights.

In contrast, we propose a novel learning based formulation to control the amount of light diffusion in portraits, without changing the overall scene illumination while softening or removing shadow and specular highlights. 

%% file: method.tex
\section{A Framework for Light Diffusion}
\label{sec:method}

In this section, we formulate the light diffusion problem, and then propose a learning based solution for in-the-wild portraits. Finally we show how our model can be applied to infer a more robust albedo from images, improving downstream applications such as relighting, face part segmentation, and normal estimation.

\subsection{Problem Formulation}

\begin{figure}
    \centering
    \includegraphics[width=\linewidth]{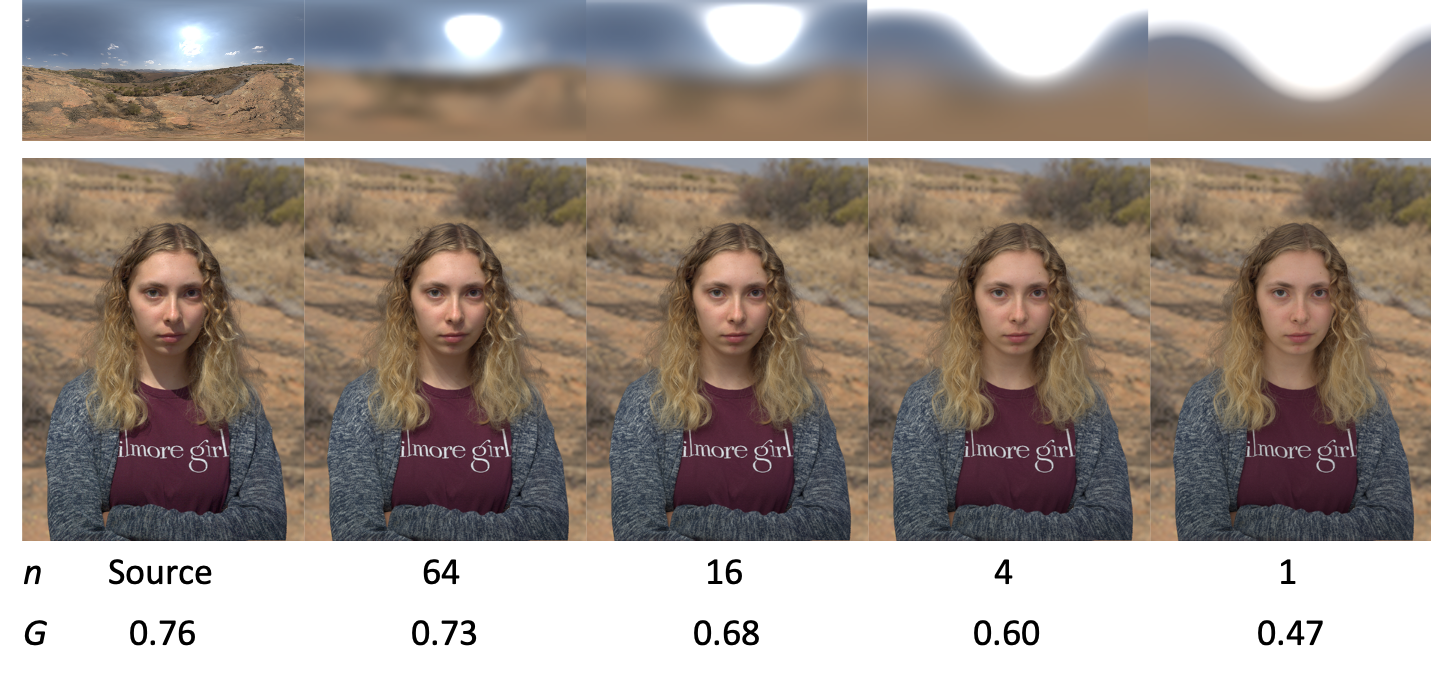}
    \caption{Illumination convolution. Shown are the original environment and relit image, followed by convolution with $\cos^n_+\theta$ with varying exponent $n$, and the resulting Gini coefficient $G$ for each diffused environment. Note the gradual reduction in light harshness while still maintaining the overall lighting tone.}
    \label{fig:diffuse_conv}
\end{figure}

We model formation of image $I$ of a subject $P$ in terms of illumination from a HDR environment map $E(\theta,\phi)$:
\begin{equation}
  I = R[P,E(\theta,\phi)]
\end{equation}
where $R[\cdot]$ renders the subject under the given HDR environment map.  We can then model light diffusion as rendering the subject under a smoothed version of the HDR environment map.  Concretely, a light-diffused image $I_d$ is formed as:
\begin{equation}
  I_d = R\left[ P,E(\theta,\phi) \conv \frac{\cos_+^n(\theta)}{\sum_{i, j}^{H, W}\cos_+^n(\theta_{i, j})}\right] \label{eqn:render-diffusion-corrected}
\end{equation}
where $\conv$ represents spherical convolution, and the incident HDR environment map is smoothed with normalized kernel $\cos_+^n(\theta) \equiv \max(0,\cos^n(\theta))$, effectively pre-smoothing the HDR environment map with the Phong specular term.  The exponent $n$ controls the amount of blur or diffusion of the lighting.  Setting $n$ to 1 leads to a diffusely lit image, and higher specular exponents result in sharper shadows and specular effects, as seen in Figure \ref{fig:diffuse_conv}. 

Our goal then is to construct a function $f$ that takes $I$ and the amount of diffusion controlled by exponent $n$ and predicts the resulting light-diffused image $I_d =  f(I, n)$. In practice, as described in section \ref{sec:parametric-diffusion-net}, we replace $n$ with a parameter $t$ that proved to be easier for the networks to learn; this new parameter is based on a novel application of the Gini coefficient~\cite{gini1912variabilita} to measure environment map diffuseness.

\subsection{Learning-based Light Diffusion}
\label{sec:model-arch}

\begin{figure*}
    \centering
    \includegraphics[width=\textwidth]{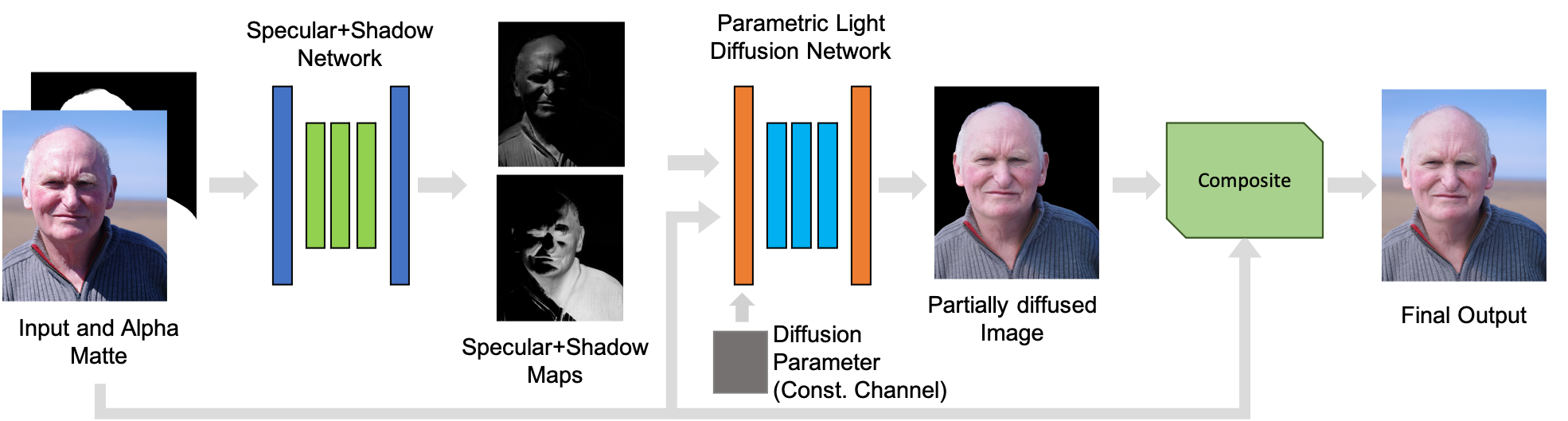}
    \caption{Architecture for parametric diffusion. Taking a portrait image with an alpha matte, the first stage predicts specular and shadow maps. The second stage uses these maps and the source image to produce an image with light diffused according to an input diffusion parameter.  The result is composited over the input image to replace the foreground subject with the newly lit version.}
    \label{fig:architecture}
\end{figure*}

We perform the light diffusion task in a deep learning framework.  We can represent the mapping $f$ as a deep network $f_\beta$:
\begin{equation}
  I_d = f_\beta(I, t)
\end{equation}
where $\beta$ represents the parameters of the network.  To supervise training, we capture subjects in a light stage and, using the OLAT images \cite{debevec00}, synthetically render each subject under an HDRI environment $E(\theta,\phi)$ and a diffused version of that environment $E(\theta,\phi) \conv \cos_+^n(\theta)$, providing training pair $I$ and $I_d$.

In practice, we obtain better results with a sequence of two networks.  The first network estimates a specular map $S$, which represents image brightening, and a shadow map $D$, which represents image darkening, both relative to a fully diffusely lit ($n=1$) image.  Concretely, we generate the fully diffused image $I_{\rm diffuse}$ as described in Equation \ref{eqn:render-diffusion-corrected} with $n = 1$ and then define the shadow $D$ and specular $S$ maps as:
\begin{eqnarray}
  S = \max(\min(1 - I_{\rm diffuse}/I,1),0) \\
  D = \max(\min(1 - I/I_{\rm diffuse},1),0)
\end{eqnarray}
Given the light stage data, it is easy to additionally synthesize $I_{\rm diffuse}$ and compute $S$ and $D$ for a given HDR environment map $E$ to supervise training of a shadow-specular network $g_{\beta_s}$:
\begin{equation}
 \{ S,D \} = g_{\beta_{s}}(I)
\end{equation}
The light diffusion network then maps the input image along with the specular and shadow maps to the final result:
\begin{equation}
  I_d = h_{\beta_d}(I,S,D,t)
\end{equation}
Note that, as we are not seeking to modify lighting of the background, we focus all the computation on the subject in the portrait.  We thus estimate a matte for the foreground subject and feed it into the networks as well; $I$ then is represented as the union of the original image and its portrait matte.  The overall framework is shown in Figure~\ref{fig:architecture}.

In addition, we can extend our framework to infer a more robust albedo than prior work, through a process of repeated light diffusion.  We now detail each of the individual components of the light diffusion and albedo estimation networks.

\subsubsection{Network details}
\label{sec:parametric-diffusion-net}
\paragraph{Specular+Shadow Network} The specular+shadow network $g_{\beta_{s}}$ is a single network that takes in the source image $I$ along with a pre-computed alpha matte \cite{pandey2021total}, as a $1024 \times 768 \times 4$ dimensional tensor. We used a U-Net \cite{unet} with 7 encoder-decoder layers and skip connections. Each layer used a $3 \times 3$ convolution followed by a Leaky ReLU activation. The number of filters for each layer is $24,  32,  64,  64, 64, 92, 128$ for the encoder, 128 for the bottleneck, and $128, 92,  64, 64, 64,  32, 24$ for the decoder. We used blur-pooling \cite{zhang2019shiftinvar} layers for down-sampling and bilinear resizing followed by a $3 \times 3$ convolution for upsampling. The final output - two single channel maps - is generated by a $3\times3$ convolution with two filters.

\paragraph{Parametric Diffusion Net}
The diffusion network $h_{\beta_d}$ takes the source image, alpha matte, specular map, shadow map, and the diffusion parameter $t$ (as a constant channel) into the Diffusion Net as a $1024 \times 768 \times 7$ tensor. The Diffusion Net is a U-Net similar to the previous U-Net, with $48,  92,  128, 256, 256, 384, 512$ encoder filters, $512$ bottleneck filters, and $384, 384, 256, 256, 128,  92, 48$ decoder filters. The larger filter count accounts for the additional difficulty of the diffusion task.


\paragraph{Diffusion parameter choice} The diffusion parameter $t$ indicates the amount of diffusion. While one can naively rely on specular exponents as a control parameter, we observed that directly using them led to poor and inconsistent results, as the perceptual change for the same specular convolution can be very varied for different HDR environments, for instance, a map with evenly distributed lighting will hardly change, whereas a map with a point light would change greatly. We hypothesize the non-linear nature of this operation is difficult for the model to learn, and so we quantified a different parameter based on a measure of `absolute' diffuseness.

\begin{figure}
    \centering
    \includegraphics[width=\linewidth]{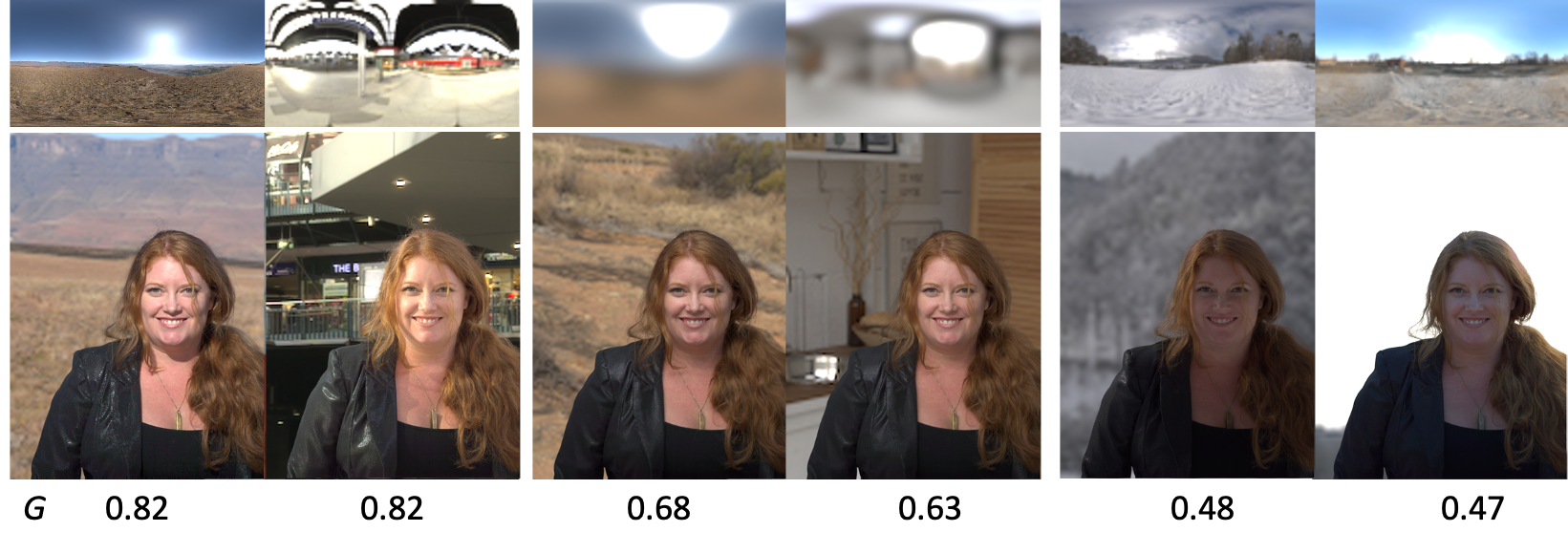}
    \caption{Gini coefficients $G$ of some HDR maps and their relit images. Similar Gini coefficients approximately yield a similar quality of lighting, allowing a consistent measure of diffusion.}
    \label{fig:gini_coefficient}
\end{figure}

To measure the absolute diffusivity of an image, we observed that the degree of diffusion is related to how evenly distributed the lighting environment is, which strongly depends on the specific scene; e.g., if all the lighting comes from a single, bright source, we will tend to have harsh shadows and strong specular effects, but if the environment has many large area lights, the image will have soft shadows and subdued specular effects. In other words, the diffusivity is related to the inequality of the lighting environment. Thus, we propose to quantify the amount of diffusion by using the \emph{Gini coefficient} \cite{gini1912variabilita} of the lighting environment, which is designed to measure inequality. Empirically, we found that the Gini coefficient gives a normalized measurement of the distribution of the light in an HDR map, as seen in Figure \ref{fig:gini_coefficient}, and thus we use it to control the amount of diffusion.


Mathematically, for a finite multiset $X \subset \mathbb{R^{+}}$, where $|X| = k$, the Gini coefficient, $G \in [0,1]$, is computed as
\begin{equation}
G = \frac{\sum_{x_i, x_j \in X} \vert x_i - x_j \vert }{2k\sum_{x_i \in X}x_i}.
\end{equation}
For a discrete HDR environment map, we compute the Gini coefficient by setting each $x_i \in X$ to be the luminance from the $i^{\text{th}}$ sample of the HDR environment map. For instance, on a discrete equirectangular projection $E(\theta, \phi)$ where $(\theta, \phi) \in [0, \pi] \times [0, 2\pi)$, the $i^{\text{th}}$ sample's light contribution is given by $E(\theta_i, \phi_i)\sin(\theta_i)$, where $\sin(\theta_i)$ compensates for higher sampling density at the poles. If we indicate the $i^{\text{th}}$ sample of $E$ by $E_i$, the coefficient is then given by
\begin{equation}
G = \frac{\sum_{i}\sum_{j} \vert E_i\sin(\theta_i) - E_j\sin(\theta_j)\vert}{2k\sum_{i}E_i\sin(\theta_i)}
\label{eqn:gini}
\end{equation}
where $i, j$ range over all samples of the equirectangular map and $k$ is the total number of samples in the map.

Finally, as an input parameter, we re-scaled this absolute measure based on each training example: $t = (G_t - G_d) / (G_s - G_d)$, where $G_t$ is the Gini coefficient for the target image, $G_d$ is the Gini coefficient for the fully diffused image (diffused with specular exponent 1), and $G_s$ is the Gini coefficient for the source image.  Parameter $t$ ranges from 0 to 1, where 0 corresponds to maximally diffuse ($G_t = G_d$) and 1 corresponds to no diffusion ($G_t = G_s)$.


\subsubsection{Albedo Estimation}
\begin{figure*}
    \centering
    \includegraphics[width=\textwidth]{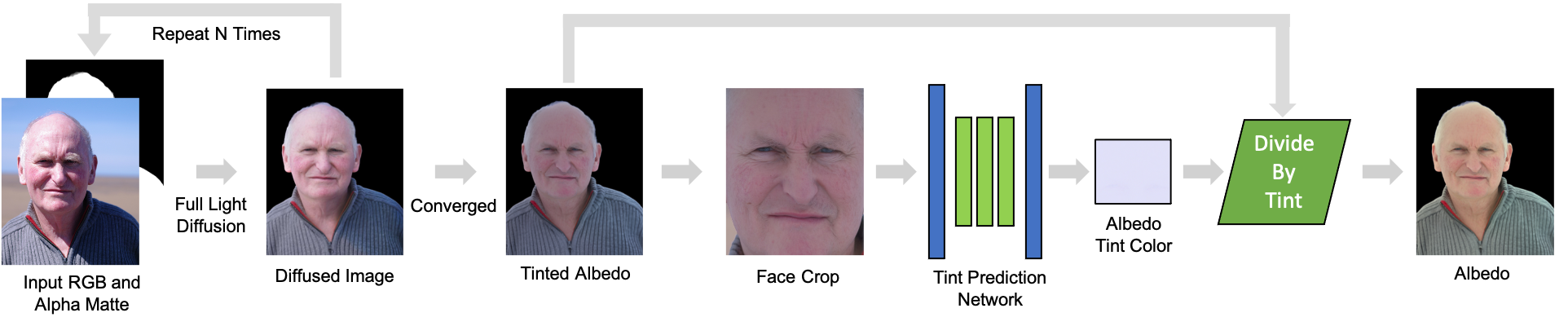}
    \caption{Architecture of our proposed extension of light diffusion to albedo prediction. We recurrently apply diffusion network $N$ times to an image, yielding an albedo map tinted by the average color of scene light. We train a model to estimate this tint, then divide it out to produce an untinted albedo image.  In this case, a warmer albedo color arises after removing the blue tint introduced by the sky illumination.}
    \label{fig:albedo-pipeline}
\end{figure*}

We observed that the primary source of errors in albedo estimation in state-of-the-art approaches like \cite{pandey2021total} arises from color and material ambiguities in clothing and is exacerbated by shadows. The albedo estimation stage tends to be the quality bottleneck in image relighting, as errors are propagated forward in such multistage pipelines. Motivated by this observation, we propose to adapt our light diffusion approach to albedo estimation (Figure~\ref{fig:albedo-pipeline}).

While the fully diffuse image (diffused with $n=1$) removes most shading effects, the approach can be pushed further to estimate an image only lit by the average color of the HDRI map, i.e., a \emph{tinted albedo}. Since the diffuse convolution operation preserves the average illumination of the HDR environment map and acts as a strong smoothing operation, repeated convolution converges to the average color of the HDR environment map. We found that iterating our diffusion network just three times (along with end-to-end training of the iteration based network) yielded good results. An alternative formulation of this problem is to pass the fully diffuse image into a separate network which estimates this tinted albedo, and we show a comparison between these two in the supplementary material. 

To remove the color tint, we crop the face -- which resides in the more constrained space of skin tone priors -- and train a CNN to estimate the RGB tint of the environment, again supervised by light stage data.  We then divide out this tint to recover the untinted albedo for the foreground.

\subsection{Data Generation and Training Details}
To train the proposed model, we require supervised pairs of input portraits $I$ and output diffused images $I_d$. Following previous work \cite{pandey2021total}, we rely on a light stage \cite{guo19,deep_relightables} to capture the full reflectance field of multiple subjects as well as their geometry. Our data generation stage consists of generation of images with varying levels of diffusion as well as the tinted and true albedo maps, to use as ground truth to train our model. 

Importantly, we also propose a synthetic shadow augmentation strategy to add external shadow with subsurface scattering effects that are not easily modeled in relit images generated in the light stage. We extend the method proposed by~\cite{zhang2020portrait} to follow the 3D geometry of the scene, by placing a virtual cylinder around the subject with a silhouette mapped to the surface.  We then project the silhouette over the 3D surface of the subject -- reconstructed from the light stage dataset -- from the strongest light in the scene followed by blurring and opacity adjustment of the resulting projected shadow map, guided by the Gini coefficient of the environment (smaller Ginis have more blur and lower shadow opacity). The resulting shadow map is used to blend between the original image and the image after removing the brightest light direction contribution.  This shadow augmentation step is key to effective light diffusion and albedo estimation.

We also augmented with subsurface scattering effects, since the light stage dataset does not include hard shadows cast by foreign objects. We implemented a heuristic approach which uses the shadow map and a skin segmentation map to inpaint a red tint around shadow edges inside detected skin regions.

For more details on our training dataset, our approach for shadow augmentation, and specifics of model training and loss functions, please refer to our supplementary material.

%% file: experiments.tex
\section{Experiments}
\label{sec:experiments}

In this section we experimentally verify the design choices of our architecture with qualitative and quantitative comparisons, and also show the effect of light diffusion on inputs for tasks like geometry estimation, semantic segmentation and portrait relighting.

\subsection{Evaluation Datasets}

We created two evaluation datasets, one from light stage images, where ground truth is available, and the other from a selection of in the wild images with harsh lighting conditions. Despite having no ground truth, qualitative results on the in-the-wild set are critical in evaluating the generalization ability of our model.

The light stage dataset consists of six subjects with diverse skin tones and genders lit by ten challenging lighting environments. The subjects as well as the HDRI maps are withheld from training and used to compute quantitative metrics against the ground truth. 

The in-the-wild dataset consists of 282 diverse portrait images in various realistic lighting conditions that highlight the usefulness of the proposed Light Diffusion. We use this set to show qualitative results for ablation studies and comparison against the state-of-the-art.

\subsection{Full Diffusion Results}

\begin{figure}[h]
    \centering
    \includegraphics[width=\linewidth]{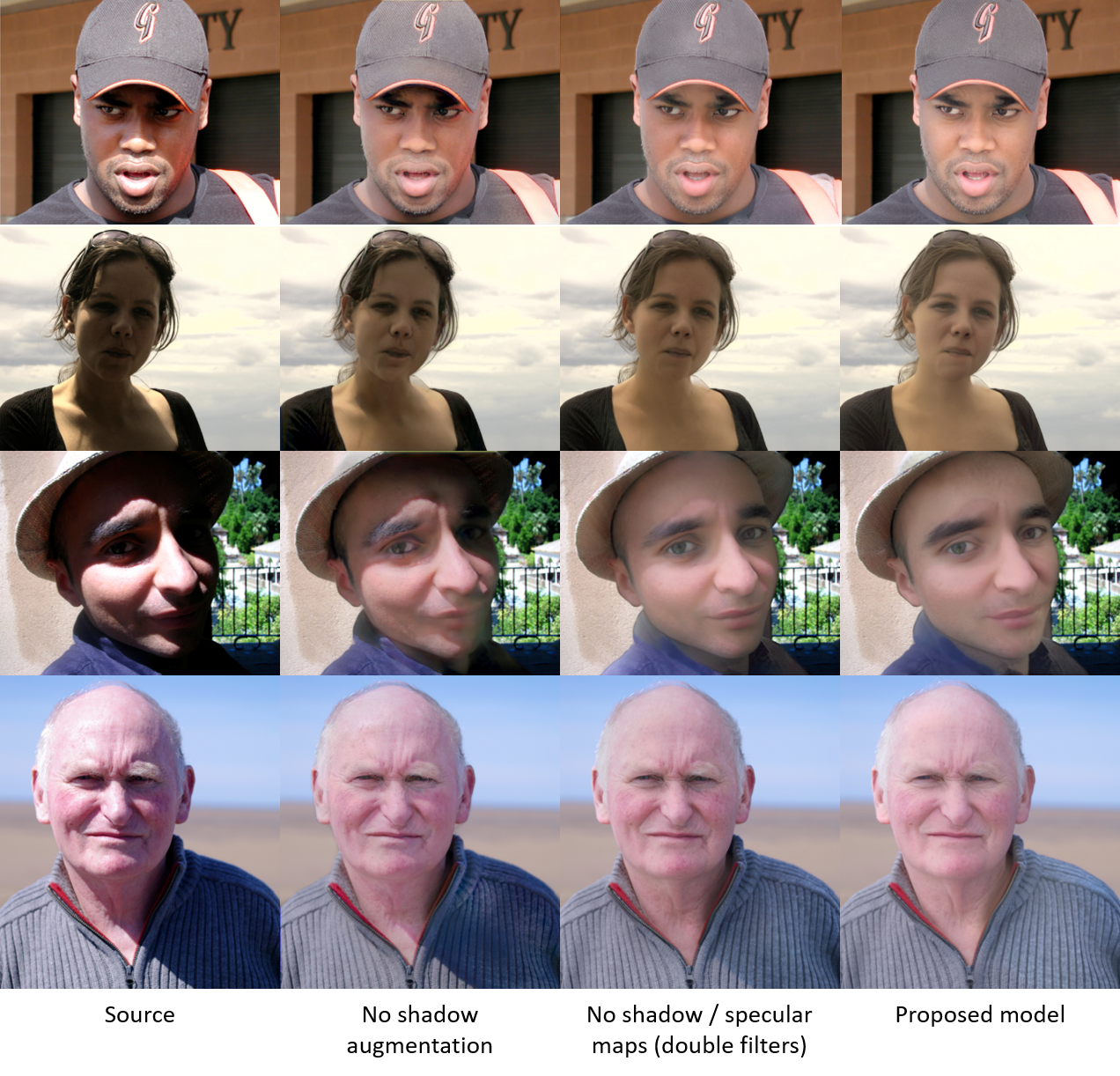}
    \caption{Fully diffuse ablation study. Left to right: source image, model with no shadow augmentation, model with twice as many filters but no shadow or specular map, proposed model.
    }
    \label{fig:ablation_diffusion}
\end{figure}

First, we trained models that predicted fully diffuse results -- the most difficult case -- to compare various design choices of the proposed algorithm. Figure~\ref{fig:ablation_diffusion} shows a comparison among three architectures. From left to right these are, (1) A model trained to predict a fully diffuse image on data without any of our proposed external shadow augmentation techniques, (2) A large model with twice as many filters but without the shadow/specular maps prediction network, and (3) Our proposed architecture. Note that the model trained without shadow augmentation does significantly worse with shadow removal, and even a larger model trained on this data struggles with shadow edges, despite having a comparable number of parameters to our proposed approach. Table~\ref{tab:ablation_diffusion} shows quantitative metrics computed on the light stage data. The large U-Net (with no shadow/specular maps) does marginally better on average metrics, however, as shown in Figure~\ref{fig:ablation_diffusion}, this does not hold on qualitative results on in-the-wild data, suggesting the increased parameter count caused overfitting to light stage images.

\begin{center}
\begin{table}
\begin{tabular}{l c c c c} 
 Model & MAE $\downarrow$ & MSE $\downarrow$ & SSIM  $\uparrow$ & LPIPS $\downarrow$ \\
 \hline
 Proposed & 0.0098 & 0.0006 & 0.9692 & 0.0340 \\ 
 No shadow aug. & 0.0123 & 0.0009 & 0.9563 & 0.0503 \\
 Large model & 0.0094 & 0.0005 & 0.9749 & 0.0304 
\end{tabular}
\caption{Quantitative metrics for fully diffuse prediction ablation study. Although the large model with no shadow/specular maps seems to do slightly better on average, results on in-the-wild data (Figure~\ref{fig:ablation_diffusion}) suggests that it substantially overfits to the light stage data.}
\label{tab:ablation_diffusion}
\end{table}
\end{center}

\subsection{In-the-Wild Applications}
In this section, we show the results of our approach on a variety of applications. As byproduct, the proposed framework can be used for multiple computational photography experiences as well as to improve downstream computer vision tasks.

\begin{figure}
    \centering
    \includegraphics[width=0.95\linewidth]{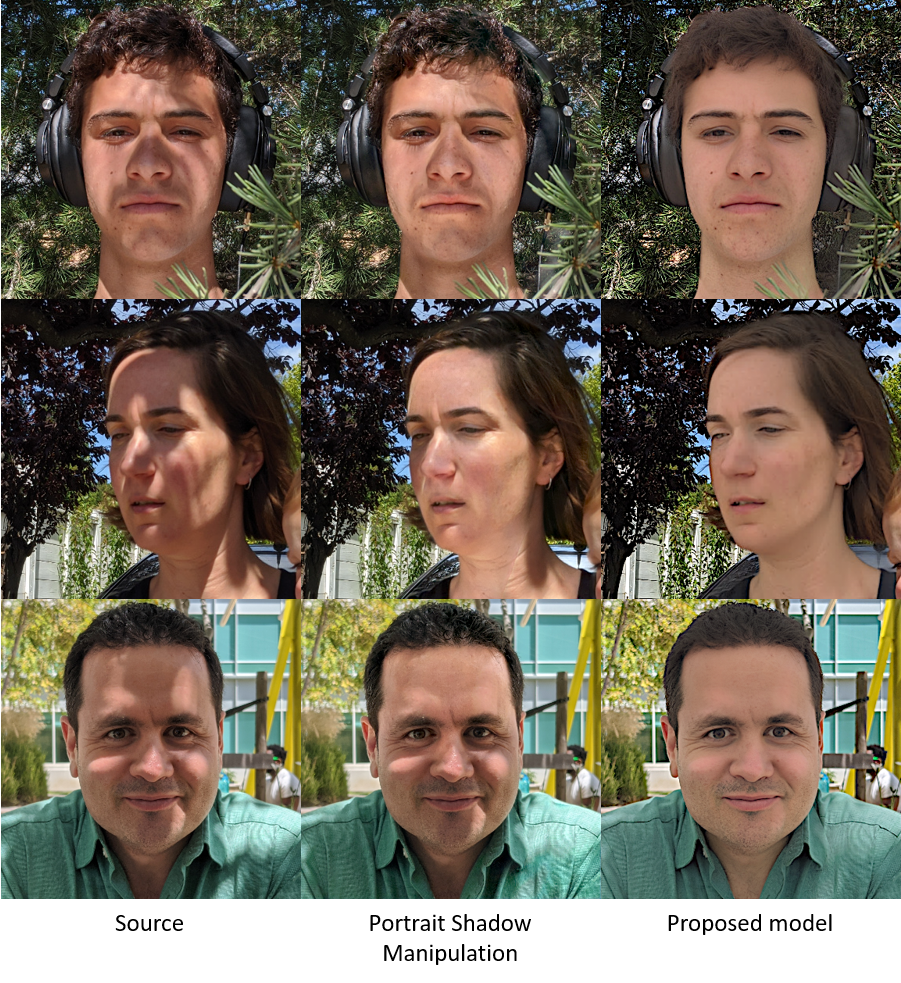}
    \caption{Comparison with \cite{zhang2020portrait}. Note how our method better softens portraits, removing shadows and specular highlights.}
    \label{fig:shadow_manip_comparison}
\end{figure}

\paragraph{Shadow Removal.}
In Figure \ref{fig:shadow_manip_comparison} we compare our fully diffuse output with that of the shadow removal approach proposed in \cite{zhang2020portrait}. Note that our model not only diffuses hard shadow edges better, but can also reduce harsh specular effects on the skin.

\begin{figure}[h]
    \centering
    \includegraphics[width=0.95\linewidth]{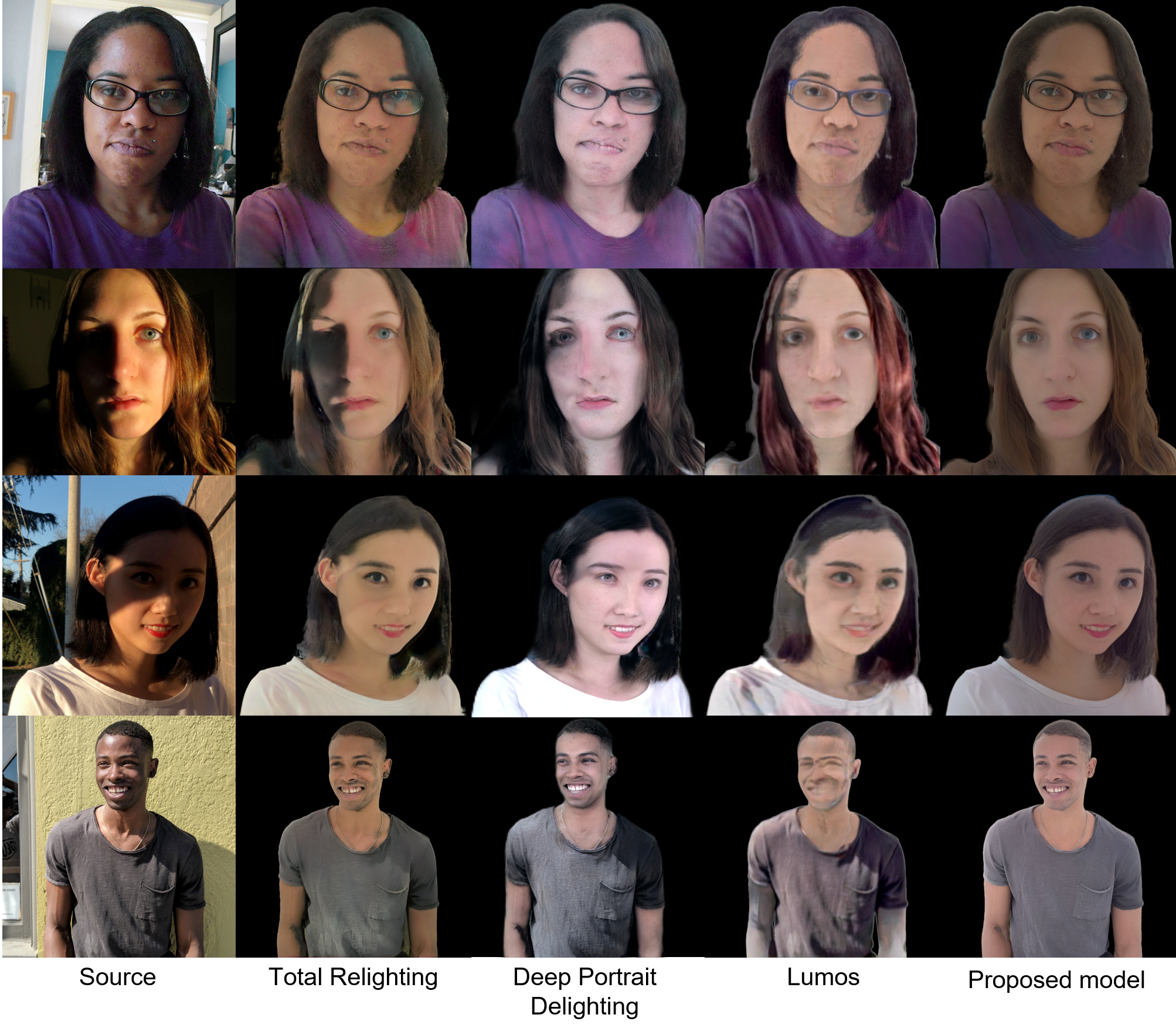}
    \caption{Albedo prediction comparisons against state of the art. Our approach has markedly better color stability, shadow removal, and skin tone preservation across subjects. From left to right: source, Total Relighting \cite{pandey2021total}, Deep Portrait Delighting \cite{Weir2022}, Lumos \cite{yeh2022lumos}, our model.}
    \label{fig:albedo_sota_comparison}
\end{figure}

\paragraph{Albedo Prediction.}
In Figure~\ref{fig:albedo_sota_comparison} we compare our approach to state-of-the-art delighting approaches~\cite{pandey2021total, Weir2022, yeh2022lumos}. Note that all previous approaches suffer from artifacts on clothing due to color ambiguities or harsh shadows, whereas our proposed algorithm correctly removes shadows and produces accurate skin tones, with no artifacts on clothing. We also provide an ablation on albedo prediction architecture choices in our supplementary text.

\begin{figure}[h]
    \centering
    \includegraphics[width=0.95\linewidth]{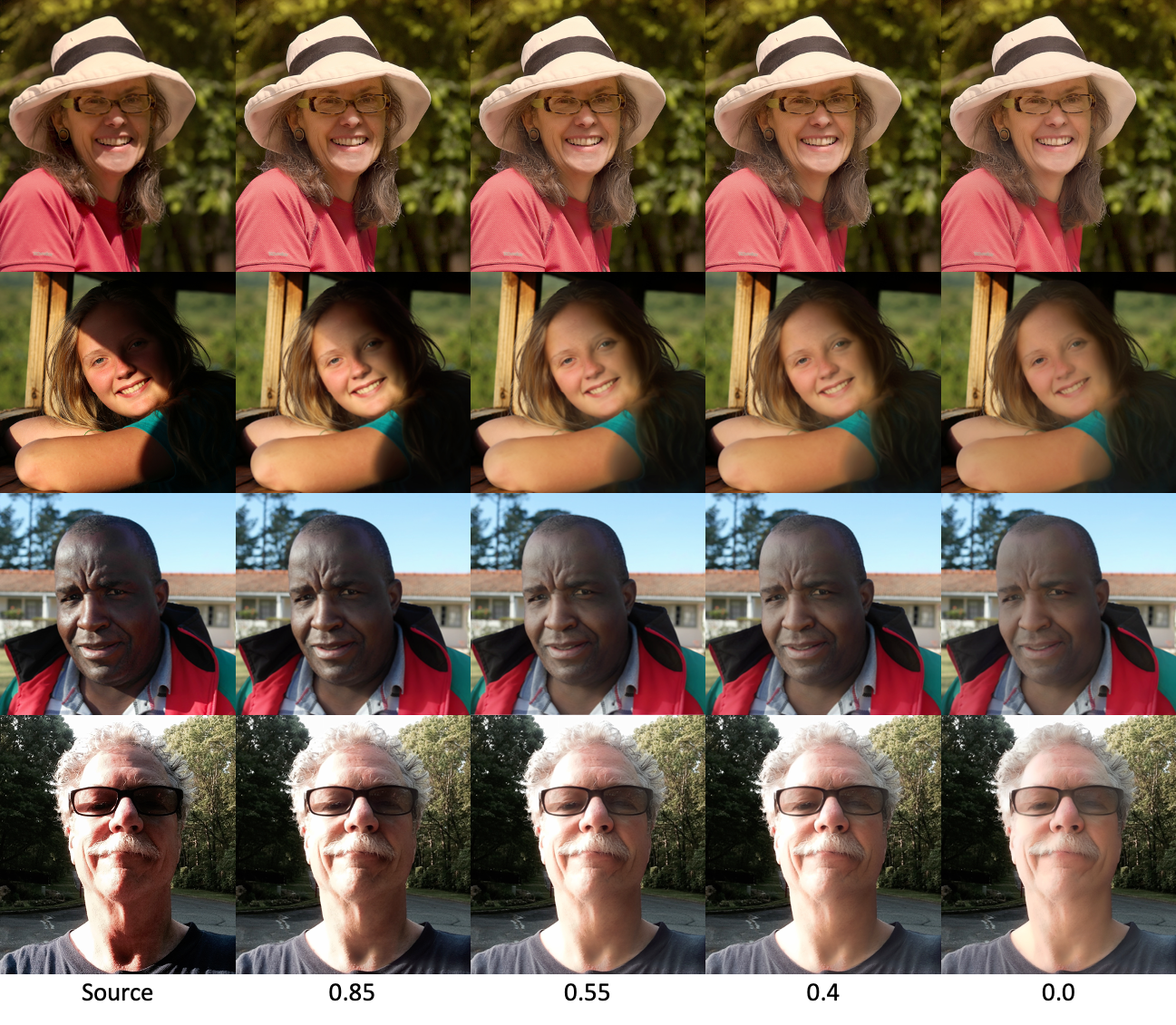}
    \caption{Parametric diffusion results. The model is able to gradually remove harsh shadows and specular effects, across skin tones, complex and deep shadows, and highly saturated images. Shown is the source image, followed by our model output at the shown diffusion parameters.}
    \label{fig:parametric_diffusion_results}
    \vspace{-15pt}
\end{figure}

\paragraph{Parametric Diffusion.}
In Figure~\ref{fig:parametric_diffusion_results} we also show the range of outputs our parametric diffusion model can produce. This is a critical feature of our proposed application, since the fully diffuse output may appear too ``flat" for compositing back into the original scene. With such control scheme, a user can choose the level of diffusion according to the level of contrast / drama they might be going for in the portrait. Note that our model can produce an aesthetically pleasing range of diffusion levels, with speculars being gradually reduced and harsh shadow edges being realistically blurred. A naive interpolation approach between the source and fully diffuse output would leave behind unnatural shadow edges. 

\begin{figure}[h]
    \centering
    \includegraphics[width=0.95\linewidth]{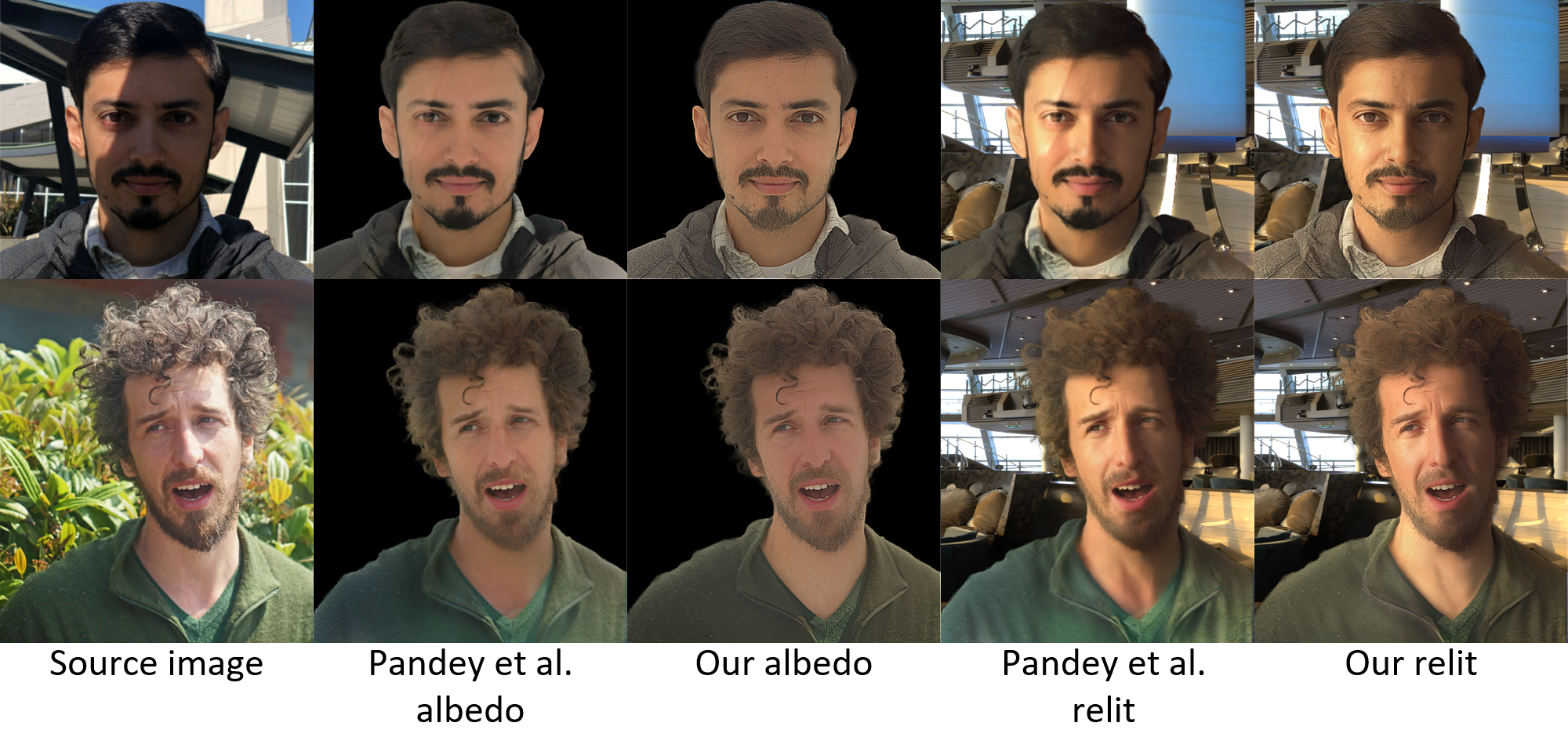}
    \caption{Example of replacing albedo prediction within Total Relighting~\cite{pandey2021total} with our albedo. From left to right: input image, albedo estimated by Total Relighting, albedo estimated by our method, image relit by Total Relighting using original albedo and image relit by Total Relighting using our albedo. Our approach is notably better at dealing with external shadows (top row) and clothing discoloration (bottom row), resulting in more realistic relighting results.}
    \label{fig:albedo_surgery}
     \vspace{-12pt}
\end{figure}

\paragraph{Portrait Relighting.}
As mentioned in the previous sections, the albedo estimation stage tends to be the bottleneck in quality for state-of-the-art portrait relighting approaches like~\cite{pandey2021total}. In Figure~\ref{fig:albedo_sota_comparison} we show that our albedo estimation approach is significantly more robust to artifacts that arise from color ambiguities and harsh shadows. In Figure~\ref{fig:albedo_surgery} we show the effect of using our estimated albedo and feeding it into the relighting module of \cite{pandey2021total}. Note that the relit result quality greatly benefits from our albedo, and no longer shows artifacts on clothing or harsh shadow regions.

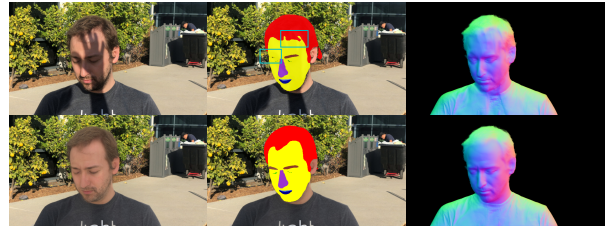
\begin{figure}[h]
    \centering
    \def\svgwidth{0.95\hsize}
    \import{_figs/parsing/}{parsing.tex}
    \caption{Light diffusion (bottom left) can improve results of state-of-the-art image processing methods, such as face parsing~\cite{Lin21} (middle, improvement in green boxes) or normal map estimation (right, removal of shadow embossing).
    }
    \label{fig:applications}
    \vspace{-12pt}
\end{figure}

\paragraph{Other Applications.}
While controlling the amount of light diffusion in a portrait is a crucial feature itself for computational photography, it can also be used as pre-processing step to simplify other computer vision tasks. The importance of relighting as a data augmentation strategy has been demonstrated in various contexts \cite{chogovadze2021controllable,tan2022volux}, and here we show that reducing the amount of unwanted shadows and specular highlights has a similar beneficial effect to downstream applications. In Figure \ref{fig:applications} we show the effect of using a diffuse image instead of the original for an off-the-shelf normal map estimator \cite{pandey2021total} and semantic segmenter for face parsing~\cite{Lin21}. In both cases, artifacts due to the external shadow are removed by using the fully diffuse input.

%% file: _figs/parsing/parsing.tex
\begingroup%
  \makeatletter%
  \providecommand\color[2][]{%
    \errmessage{(Inkscape) Color is used for the text in Inkscape, but the package 'color.sty' is not loaded}%
    \renewcommand\color[2][]{}%
  }%
  \providecommand\transparent[1]{%
    \errmessage{(Inkscape) Transparency is used (non-zero) for the text in Inkscape, but the package 'transparent.sty' is not loaded}%
    \renewcommand\transparent[1]{}%
  }%
  \providecommand\rotatebox[2]{#2}%
  \newcommand*\fsize{\dimexpr\f@size pt\relax}%
  \newcommand*\lineheight[1]{\fontsize{\fsize}{#1\fsize}\selectfont}%
  \ifx\svgwidth\undefined%
    \setlength{\unitlength}{1733.38275146bp}%
    \ifx\svgscale\undefined%
      \relax%
    \else%
      \setlength{\unitlength}{\unitlength * \real{\svgscale}}%
    \fi%
  \else%
    \setlength{\unitlength}{\svgwidth}%
  \fi%
  \global\let\svgwidth\undefined%
  \global\let\svgscale\undefined%
  \makeatother%
  \begin{picture}(1,0.37342685)%
    \lineheight{1}%
    \setlength\tabcolsep{0pt}%
    \put(0,0){\includegraphics[width=\unitlength,page=1]{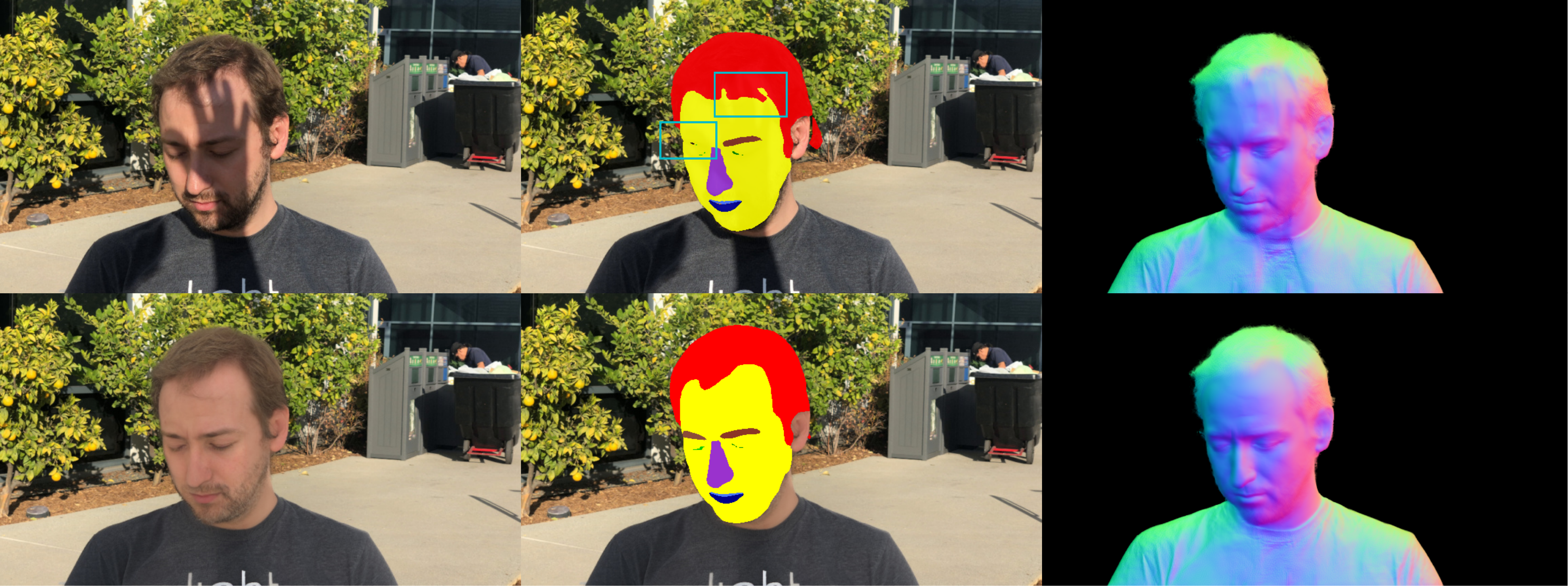}}%
  \end{picture}%
\endgroup%

%% file: conclusion.tex
\section{Discussion}
\label{sec:conclusion}
We proposed a complete system for \emph{light diffusion}, a novel method to control the lighting in portrait photography by reducing harsh shadows and specular effects. Our approach can be used directly to improve photographs and can aid numerous downstream tasks. In particular, we have shown that light diffusion generalizes well to albedo prediction, greatly improving on the state of the art. We have also shown that geometry estimation and semantic segmentation is improved, and we expect that the process should improve many other downstream portrait-based vision tasks.

\begin{figure}[h]
    \centering
    \includegraphics[width=0.95\linewidth]{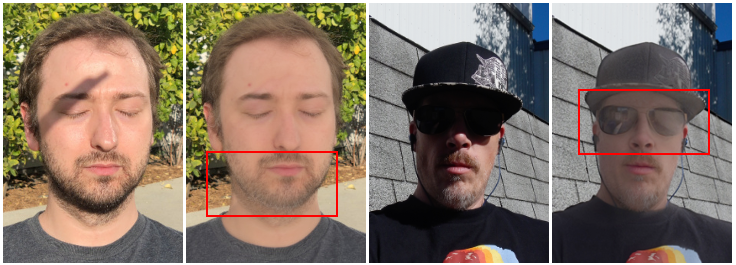}
    \caption{Limitations of our approach. After full diffusion, the model can sometimes lighten facial hair. The model also has trouble with dark sunglasses, tending to inpaint them with skin shades.}
    \label{fig:limitations}
    \vspace{-15pt}
\end{figure}

\paragraph{Limitations.} Despite vastly increasing the domain of materials that can have lighting adjustments, our model has some limitations, as shown in Figure \ref{fig:limitations}. In particular, we notice that dark facial hair tends to be lightened, perhaps due to its resemblance to a shadow region. In addition, our model has trouble with dark sunglasses, tending to add skin tones to them. Other limitations include over-blurring excessively diffused images where fine details should be synthesized instead, failure to remove very strong specularities and occasional confusion of objects for shadows.

\paragraph{Fairness.} Our results have shown that our proposed approach works well across a variety of skin tones. To validate this, we also ran a detailed fairness study to analyze results across different Fitzpatrick skin tones. Please refer to our supplementary material for details.

%% file: main.bbl
\begin{thebibliography}{10}\itemsep=-1pt

\bibitem{Abdal21}
Rameen Abdal, Peihao Zhu, Niloy~J. Mitra, and Peter Wonka.
\newblock Styleflow: Attribute-conditioned exploration of stylegan-generated
  images using conditional continuous normalizing flows.
\newblock {\em ACM Transactions on Graphics}, 2021.

\bibitem{Abdelrehim07}
Abdelrehim Ahmed and Aly Farag.
\newblock A new statistical model combining shape and spherical harmonics
  illumination for face reconstruction.
\newblock In {\em Advances in Visual Computing}, 2007.

\bibitem{BarronM15}
Jonathan~T. Barron and Jitendra Malik.
\newblock {Shape, Illumination, and Reflectance From Shading}.
\newblock {\em {IEEE} Transactions on Pattern Analysis and Machine
  Intelligence}, 2015.

\bibitem{blanz99}
Volker Blanz and Thomas Vetter.
\newblock A morphable model for the synthesis of 3d faces.
\newblock In {\em SIGGRAPH}, 1999.

\bibitem{chogovadze2021controllable}
George Chogovadze, Rémi Pautrat, and Marc Pollefeys.
\newblock Controllable data augmentation through deep relighting, 2021.

\bibitem{debevec00}
Paul Debevec, Tim Hawkins, Chris Tchou, Haarm-Pieter Duiker, Westley Sarokin,
  and Mark Sagar.
\newblock Acquiring the reflectance field of a human face.
\newblock SIGGRAPH '00, page 145–156, USA, 2000. ACM Press/Addison-Wesley
  Publishing Co.

\bibitem{deng2020disentangled}
Yu Deng, Jiaolong Yang, Dong Chen, Fang Wen, and Xin Tong.
\newblock Disentangled and controllable face image generation via 3d
  imitative-contrastive learning.
\newblock In {\em IEEE Computer Vision and Pattern Recognition}, 2020.

\bibitem{gini1912variabilita}
Corrado Gini.
\newblock {\em Variabilit{\`a} e mutabilit{\`a}: contributo allo studio delle
  distribuzioni e delle relazioni statistiche.[Fasc. I.]}.
\newblock Tipogr. di P. Cuppini, 1912.

\bibitem{grey04}
Christopher Grey.
\newblock Master lighting guide for portrait photographers.
\newblock In {\em Amherst Media}, 2004.

\bibitem{guo19}
Kaiwen Guo, Peter Lincoln, Philip Davidson, Jay Busch, Xueming Yu, Matt Whalen,
  Geoff Harvey, Sergio Orts-Escolano, Rohit Pandey, Jason Dourgarian, Danhang
  Tang, Anastasia Tkach, Adarsh Kowdle, Emily Cooper, Mingsong Dou, Sean
  Fanello, Graham Fyffe, Christoph Rhemann, Jonathan Taylor, Paul Debevec, and
  Shahram Izadi.
\newblock The relightables: Volumetric performance capture of humans with
  realistic relighting.
\newblock {\em ACM Trans. Graph.}, 38(6), nov 2019.

\bibitem{inoue21}
Naoto Inoue and Toshihiko Yamasaki.
\newblock Learning from synthetic shadows for shadow detection and removal.
\newblock {\em IEEE Transactions on Circuits and Systems for Video Technology},
  2021.

\bibitem{ji2022relight}
Chaonan Ji, Tao Yu, Kaiwen Guo, Jingxin Liu, and Yebin Liu.
\newblock Geometry-aware single-image full-body human relighting.
\newblock 2022.

\bibitem{Kanamori:2018}
Yoshihiro Kanamori and Yuki Endo.
\newblock {Relighting Humans: Occlusion-Aware Inverse Rendering for Full-Body
  Human Images}.
\newblock {\em ACM Transactions Graphics (Proc. SIGGRAPH Asia)}, 2018.

\bibitem{LeGendre19}
Chloe LeGendre, Wan{-}Chun Ma, Graham Fyffe, John Flynn, Laurent Charbonnel,
  Jay Busch, and Paul~E. Debevec.
\newblock Deeplight: Learning illumination for unconstrained mobile mixed
  reality.
\newblock {\em CVPR}, 2019.

\bibitem{LeGendre20}
Chloe LeGendre, Wan-Chun Ma, Rohit Pandey, Sean Fanello, Christoph Rhemann,
  Jason Dourgarian, Jay Busch, and Paul Debevec.
\newblock Learning illumination from diverse portraits.
\newblock In {\em SIGGRAPH Asia 2020 Technical Communications}, 2020.

\bibitem{Lin21}
Yiming Lin, Jie Shen, Yujiang Wang, and Maja Pantic.
\newblock Roi tanh-polar transformer network for face parsing in the wild.
\newblock {\em Image and Vision Computing}, 112:104190, 2021.

\bibitem{mallikarjun2021}
B~R Mallikarjun, Ayush Tewari, Abdallah Dib, Tim Weyrich, Bernd Bickel,
  Hans-Peter Seidel, Hanspeter Pfister, Wojciech Matusik, Louis Chevallier,
  Mohamed Elgharib, et~al.
\newblock Photoapp: Photorealistic appearance editing of head portraits.
\newblock {\em ACM Transactions on Graphics}, 40(4):1--16, 2021.

\bibitem{meka:2017}
Abhimitra Meka, Gereon Fox, Michael Zollh{\"o}fer, Christian Richardt, and
  Christian Theobalt.
\newblock {Live User-Guided Intrinsic Video for Static Scene}.
\newblock {\em IEEE Transactions on Visualization and Computer Graphics}, 2017.

\bibitem{Meka:2018}
Abhimitra Meka, Maxim Maximov, Michael Zollhoefer, Avishek Chatterjee,
  Hans-Peter Seidel, Christian Richardt, and Christian Theobalt.
\newblock {LIME: Live Intrinsic Material Estimation}.
\newblock In {\em Proc. Computer Vision and Pattern Recognition}, 2018.

\bibitem{deep_relightables}
Abhimitra Meka, Rohit Pandey, Christian H\"{a}ne, Sergio Orts-Escolano, Peter
  Barnum, Philip David-Son, Daniel Erickson, Yinda Zhang, Jonathan Taylor,
  Sofien Bouaziz, Chloe LeGendre, Wan-Chun Ma, Ryan Overbeck, Thabo Beeler,
  Paul Debevec, Shahram Izadi, Christian Theobalt, Christoph Rhemann, and Sean
  Fanello.
\newblock Deep relightable textures: Volumetric performance capture with neural
  rendering.
\newblock {\em ACM Transactions on Graphics}, 2020.

\bibitem{nestmeyer2020learning}
Thomas Nestmeyer, Jean-François Lalonde, Iain Matthews, and Andreas~M.
  Lehrmann.
\newblock Learning physics-guided face relighting under directional light.
\newblock In {\em CVPR}, 2020.

\bibitem{pan2021shadegan}
Xingang Pan, Xudong Xu, Chen~Change Loy, Christian Theobalt, and Bo Dai.
\newblock A shading-guided generative implicit model for shape-accurate
  3d-aware image synthesis.
\newblock In {\em NeurIPS}, 2021.

\bibitem{pandey2021total}
Rohit Pandey, Sergio~Orts Escolano, Chloe Legendre, Christian Haene, Sofien
  Bouaziz, Christoph Rhemann, Paul Debevec, and Sean Fanello.
\newblock Total relighting: learning to relight portraits for background
  replacement.
\newblock {\em ACM Transactions on Graphics (TOG)}, 40(4):1--21, 2021.

\bibitem{Ren:2015}
Peiran Ren, Yue Dong, Stephen Lin, Xin Tong, and Baining Guo.
\newblock {Image Based Relighting Using Neural Networks}.
\newblock {\em ACM Transactions on Graphics}, 2015.

\bibitem{unet}
Olaf Ronneberger, Philipp Fischer, and Thomas Brox.
\newblock U-net: Convolutional networks for biomedical image segmentation.
\newblock In Nassir Navab, Joachim Hornegger, William~M. Wells, and
  Alejandro~F. Frangi, editors, {\em Medical Image Computing and
  Computer-Assisted Intervention -- MICCAI 2015}, pages 234--241, Cham, 2015.
  Springer International Publishing.

\bibitem{shu2017portrait}
Zhixin Shu, Sunil Hadap, Eli Shechtman, Kalyan Sunkavalli, Sylvain Paris, and
  Dimitris Samaras.
\newblock Portrait lighting transfer using a mass transport approach.
\newblock {\em ACM Transactions on Graphics (TOG)}, 36(4):1, 2017.

\bibitem{sun2019single}
Tiancheng Sun, Jonathan~T Barron, Yun-Ta Tsai, Zexiang Xu, Xueming Yu, Graham
  Fyffe, Christoph Rhemann, Jay Busch, Paul Debevec, and Ravi Ramamoorthi.
\newblock Single image portrait relighting.
\newblock {\em ACM Transactions on Graphics (TOG)}, 38(4):79, 2019.

\bibitem{tan2022volux}
Feitong Tan, Sean Fanello, Abhimitra Meka, Sergio Orts-Escolano, Danhang Tang,
  Rohit Pandey, Jonathan Taylor, Ping Tan, and Yinda Zhang.
\newblock Volux-gan: A generative model for 3d face synthesis with hdri
  relighting.
\newblock {\em ACM SIGGRAPH}, 2022.

\bibitem{Wang20}
Zhibo Wang, Xin Yu, Ming Lu, Quan Wang, Chen Qian, and Feng Xu.
\newblock Single image portrait relighting via explicit multiple reflectance
  channel modeling.
\newblock {\em ACM SIGGRAPH Asia and Transactions on Graphics}, 2020.

\bibitem{Weir2022}
Joshua Weir, Junhong Zhao, Andrew Chalmers, and Taehyun Rhee.
\newblock Deep portrait delighting.
\newblock {\em ECCV}, 2022.

\bibitem{xu2018deep}
Zexiang Xu, Kalyan Sunkavalli, Sunil Hadap, and Ravi Ramamoorthi.
\newblock Deep image-based relighting from optimal sparse samples.
\newblock {\em ACM Transactions on Graphics}, 2018.

\bibitem{yang22}
Xingchao Yang and Takafumi Taketomi.
\newblock {BareSkinNet: De-makeup and De-lighting via 3D Face Reconstruction}.
\newblock {\em Computer Graphics Forum}, 2022.

\bibitem{yeh2022learning}
Yu-Ying Yeh, Koki Nagano, Sameh Khamis, Jan Kautz, Ming-Yu Liu, and Ting-Chun
  Wang.
\newblock Learning to relight portrait images via a virtual light stage and
  synthetic-to-real adaptation.
\newblock {\em ACM Transactions on Graphics (TOG)}, 2022.

\bibitem{yeh2022lumos}
Yu-Ying Yeh, Koki Nagano, Sameh Khamis, Jan Kautz, Ming-Yu Liu, and Ting-Chun
  Wang.
\newblock Learning to relight portrait images via a virtual light stage and
  synthetic-to-real adaptation.
\newblock {\em ACM Transactions on Graphics (TOG)}, 2022.

\bibitem{zhang21}
Longwen Zhang, Qixuan Zhang, Minye Wu, Jingyi Yu, and Lan Xu.
\newblock Neural video portrait relighting in real-time via consistency
  modeling.
\newblock {\em CoRR}, 2021.

\bibitem{zhang2019shiftinvar}
Richard Zhang.
\newblock Making convolutional networks shift-invariant again.
\newblock In {\em ICML}, 2019.

\bibitem{zhang2020portrait}
Xuaner Zhang, Jonathan~T. Barron, Yun-Ta Tsai, Rohit Pandey, Xiuming Zhang, Ren
  Ng, and David~E. Jacobs.
\newblock Portrait shadow manipulation.
\newblock volume~39, 2020.

\bibitem{zhou2019portraitrelighting}
Hao Zhou, Sunil Hadap, Kalyan Sunkavalli, and David Jacobs.
\newblock Deep single image portrait relighting.
\newblock In {\em ICCV}, 2019.

\end{thebibliography}
